%% file: neurips_2026.tex
\documentclass{article}

\PassOptionsToPackage{numbers, compress}{natbib}

\usepackage[preprint]{neurips_2026}

\usepackage[utf8]{inputenc} 
\usepackage[T1]{fontenc}    
\usepackage{hyperref}       
\usepackage{url}            
\usepackage{booktabs}       
\usepackage{amsfonts}       
\usepackage{nicefrac}       
\usepackage{microtype}      
\usepackage{xcolor}         
\usepackage{amsmath}
\usepackage{amssymb}
\usepackage{multirow}
\usepackage{graphicx}

\title{Beyond Myopic World Models: Long-Horizon End-to-End Training for Direct Future Prediction}

\author{%
  Xinyi Li \quad Zaishuo Xia \quad Chenjie Hao \quad Yubei Chen \\
  University of California, Davis \\
  \texttt{\{kxyli, zsxia, cjhao, ybchen\}@ucdavis.edu} \\
}

\begin{document}

\maketitle

\begin{abstract}
World models are expected to support imagination over extended temporal horizons, yet most are still trained through local few-step prediction objectives and deployed by recursively rolling out their own predictions. This creates a fundamental mismatch: few-step losses optimize local transition fidelity, while long-horizon prediction depends on how errors and gradients propagate through the entire trajectory. As a result, transitions with different downstream influence on the endpoint are treated uniformly during training, and small local errors are amplified through recursive inference. We argue that long-horizon accuracy is better achieved by optimizing directly, through an end-to-end endpoint prediction objective. To instantiate this paradigm, we introduce the \textbf{Direct Prediction World Model (DPWM)}, a non-recursive architecture that compresses an action sequence of arbitrary length into a single embedding and predicts the endpoint observation in a single forward pass. This design avoids recurrent rollout in both prediction and gradient propagation, making long-horizon end-to-end training practical at horizons where unrolled autoregressive training becomes unstable. Empirically, DPWM substantially improves long-horizon endpoint prediction over recursive world-model baselines on continuous-control and pixel-based benchmarks, with larger gains as the prediction horizon increases. We further show that recurrent baselines benefit similarly when retrained with the same long-horizon endpoint objective, supporting our central claim that the training objective, rather than the particular backbone choice, is the main driver of long-horizon prediction accuracy. Our results suggest that world models can benefit from being trained and evaluated at the temporal scales where they are ultimately used, shifting the focus from local transition modeling toward long-horizon predictive accuracy.
\end{abstract}

\input{sections/intro}
\input{sections/method}

\input{sections/experiments}
\input{sections/conclusion}

\newpage

\clearpage
\bibliographystyle{plainnat}
\bibliography{main}


\input{sections/appendix}



\end{document}

%% file: sections/intro.tex
\section{Introduction}

\begin{figure}
  \centering
  \includegraphics[width=0.9\linewidth]{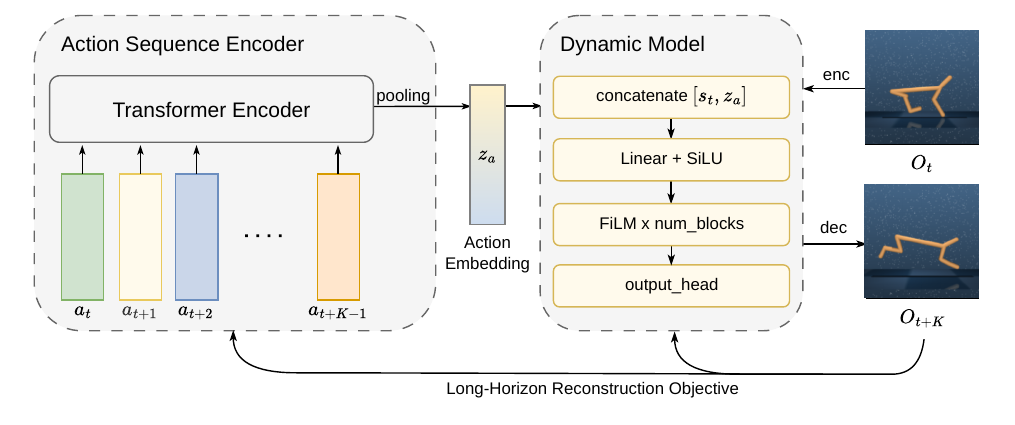}
  \caption{DPWM architecture. \textbf{Action Sequence Encoder} compresses the variable-length action sequence into a single embedding, which conditions a \textbf{Dynamic Model} that directly predicts the endpoint observation $\hat{o}_{t+K}$ from the initial observation $o_t$. Modules are trained end-to-end under a long-horizon reconstruction objective on the endpoint, without generating intermediate observations.}
\end{figure}

World models~\citep{ha2018world, lecun2022path, sutton1991dyna} provide a compact predictive interface between past experience and future decision making. In model-based reinforcement learning and embodied control, they are expected to support imagination over extended temporal horizons: an agent must evaluate the consequence of an action sequence, compare alternative plans, or reason about delayed outcomes many steps into the future~\citep{sutton1991dyna, hafner2019planet, hafner2019dreamerv1, hafner2020dreamerv2, hafner2025dreamerv3, henaff2017model, alonso2024diamond, Hao_2025_mosim}.
Long-horizon prediction is also a useful test of whether a model has captured the underlying dynamics~\citep{Vafa2025WhatHA,Liu2026FromKT,li2025smallworld, xia2026cloningdeterministicworldscritical,Finn2016UnsupervisedLF}, since errors in transition structure are often only revealed after the model is rolled forward over many steps.
However, many widely used world models are trained through local transition objectives. They learn to predict the next observation, or a short segment of future observations, and long-horizon predictions are obtained at inference time by recursively feeding the model's own predictions back into itself~\citep{hafner2019planet,hafner2019dreamerv1,micheli2023iris,alonso2024diamond}.

This creates a mismatch between how world models are trained and how they are used. A one-step objective optimizes local transition fidelity under ground-truth inputs, but long-horizon prediction depends on the entire composed trajectory under model-generated inputs: small errors drift the rollout away from the training distribution and amplify through recursive inference~\citep{bengio2015scheduledsamplingsequenceprediction, lamb2016professorforcing, asadi2019combatingcompoundingerrorproblemmultistep, talvitie2017selfcorrectingmodelsmodelbasedreinforcement}. 
More fundamentally, transitions are weighted uniformly by the local loss even though their downstream influence on the endpoint differs substantially with the dynamics and horizon, so improving one-step accuracy does not necessarily optimize the prediction an agent ultimately uses after tens or hundreds of imagined steps~\citep{asadi2019combatingcompoundingerrorproblemmultistep,benechehab2024multitimestep}. 
A natural alternative is to attach the loss directly to the endpoint induced by an initial observation and an action sequence, so that the optimization signal is tied to the long-horizon prediction target rather than to a collection of local transition errors.
However, extending this approach to long horizons introduces practical obstacles: training an autoregressive model with an endpoint loss requires unrolling the model across the full prediction horizon, increasing memory cost and optimization difficulty as the horizon grows~\citep{Pascanu2012OnTD,lamb2016professorforcing}.
Compared to short-horizon and recursively rolled-out world modeling, long-horizon end-to-end endpoint supervision remains relatively underexplored.

To study this training paradigm directly, we introduce the \textbf{Direct Prediction World Model (DPWM)}, a non-recursive world model that predicts the endpoint observation from the initial observation and the entire variable-length action sequence in a single forward pass. 
DPWM first compresses the action sequence into a single embedding and then conditions a dynamics module on both the initial observation representation and the action embedding to predict the endpoint. 
By predicting the endpoint directly rather than rolling out intermediate observations, DPWM avoids feeding prediction errors back into later prediction steps, reducing one source of compounding rollout error.
The same property also makes long-horizon end-to-end training more computationally practical because the endpoint loss is applied without recursively unrolling the model through the extended horizon.

Empirically, DPWM substantially improves endpoint prediction accuracy on continuous-control tasks in DeepMind Control Suite~\cite{tunyasuvunakool2020}, with the advantage becoming more pronounced as the evaluation horizon increases. Increasing the maximum training horizon further improves stability and generalization at long evaluation horizons, including evaluation beyond the training horizon.

The central claim of this paper, however, is not that a particular sequence encoder or backbone is inherently superior. Instead, we argue that long-horizon prediction accuracy is strongly shaped by the training objective and prediction paradigm; the architecture of DPWM serves as a practical mechanism for making this paradigm feasible at large horizons. To separate objective from architecture, we retrain autoregressive any-step baselines~\citep{admpo, adm2} under the same long-horizon endpoint objective and compare them against their original short-horizon counterparts.
We find that these baselines benefit significantly from the same endpoint supervision, supporting the view that the training paradigm, rather than the specific backbone alone, is a major driver of long-horizon prediction accuracy. 
These results suggest that world models can benefit from being trained and evaluated at the temporal scales at which they are ultimately used, rather than assuming that accurate short-horizon prediction will translate to accurate long-horizon prediction.

In summary, our main contributions are as follows:
\begin{itemize}
    \item We formulate long-horizon world modeling as direct endpoint prediction, and analyze how this paradigm and the choice of training horizon shape long-horizon prediction accuracy.
    \item We propose DPWM, a non-recursive architecture that makes this paradigm practical at extended horizons, and achieves substantial gains in endpoint prediction accuracy on continuous-control and pixel-based benchmarks. 
\end{itemize}

%% file: sections/method.tex
\section{Method}

\subsection{Direct Prediction Formulation}
\label{sec:formulation}

\textbf{Markov decision process.}
We consider a deterministic Markov decision process\footnote{We present the formulation for deterministic dynamics for clarity; in stochastic environments, the same direct-prediction view applies to modeling the conditional distribution or conditional mean of $s_{t+K}$ given $(s_t,a_{t:t+K-1})$.} $\mathcal{M}=(\mathcal{S},\mathcal{A},f)$, where $\mathcal{S}$ and $\mathcal{A}$ denote the state and action spaces, and $f:\mathcal{S}\times\mathcal{A}\to\mathcal{S}$ is the deterministic transition function. Given a state $s_t$ and an action sequence $a_{t:t+K-1}:=(a_t,\ldots,a_{t+K-1})$, the $K$-step deterministic transition map $\mathcal{S}\times\mathcal{A}^{K}\to\mathcal{S}$ induced by $f$ is the composition:
\begin{equation}
    f^{(K)}(s_t,a_{t:t+K-1})
    :=
    f(\cdots f(f(s_t,a_t),a_{t+1})\cdots,a_{t+K-1}).
\end{equation}
Although $f^{(K)}$ is by construction a $K$-fold composition of $f$, its approximator need not be recursive.

\textbf{Two paradigms.}
A standard one-step autoregressive world model learns an approximation $\hat{f}_\theta$ to $f^{(1)}=f$ and produces long-horizon predictions by recursively applying $\hat{f}_\theta$ to its own predicted states. We instead train a model $g_\theta : \mathcal{S} \times \mathcal{A}^K \to \mathcal{S}$ to approximate $f^{(K)}$ directly for variable horizons $K \in \{1, \ldots, K_{\max}\}$: rather than predicting the intermediate states $s_{t+1}, \ldots, s_{t+K-1}$, the model predicts the endpoint $s_{t+K}$ in a single forward pass.

The two paradigms differ in how prediction errors scale with $K$. Under standard $L$-Lipschitz assumptions on $f$ and a one-step approximation error $\epsilon$ for $\hat{f}_\theta$, recursive rollout produces an endpoint error bounded by $\epsilon \sum_{i=0}^{K-1} L^i$, which grows exponentially in $K$ whenever $L > 1$~\citep{asadi2018lipschitzcontinuitymodelbasedreinforcement}. This amplification is not a property of the modeling error itself, but of the recursive structure in which each step re-applies $\hat{f}_\theta$ to a perturbed input. For tasks with locally expansive dynamics (e.g., chaotic systems, contact-rich manipulation), $L > 1$ is the norm rather than the exception. Direct prediction, by approximating $f^{(K)}$ as a single mapping, incurs an endpoint error bounded only by $\delta_K := \sup \|g_\theta - f^{(K)}\|$, with no recursive amplification. A complete derivation is given in App.~\ref{app:lipschitz}.

This does not eliminate the difficulty of long-horizon modeling; it relocates it. Recursive rollout pays a horizon-dependent \emph{inference-time} cost, while direct prediction pays a \emph{modeling} cost: as $K$ grows, the target map $f^{(K)}$ becomes harder to approximate and $\delta_K$ may grow with $K$. The substantive question is therefore empirical: how low can $\delta_K$ be driven in practice, and is this enough to make direct prediction favorable at long horizons? We answer this experimentally in Sec.~\ref{sec:experiments}. The same Lipschitz factor $L$ reappears in our analysis of training-time gradient misalignment in Sec.~\ref{sec:training}.

\textbf{Instantiation.}
We instantiate the direct prediction model $g_\theta$ defined above as a composition of four learned modules. An observation encoder $E$ maps the raw observation $o_t$ to a latent representation $s_t = E(o_t)$. An action encoder $\phi$ compresses the action sequence into a fixed-dimensional embedding $z_a = \phi(a_{t:t+K-1})$. A dynamics module $g$ produces the predicted endpoint latent $\hat{s}_{t+K} = g(s_t, z_a)$. A decoder $D$ maps the predicted latent back to observation space, $\hat{o}_{t+K} = D(\hat{s}_{t+K})$. Together,
\begin{equation}
\hat{o}_{t+K} \;=\; D\!\big(\,g\big(E(o_t),\,\phi(a_{t:t+K-1})\big)\,\big),
\label{eq:pipeline}
\end{equation}
which we view as the operational form of $g_\theta$ in this paper. The next subsection describes how this composition is trained; concrete instantiations of the four modules are deferred to Sec.~\ref{sec:model_design}.

\subsection{Long-Horizon End-to-End Training}
\label{sec:training}

We train the full pipeline $(E, \phi, g, D)$ end-to-end to predict the ground-truth endpoint observation $o_{t+K}$. For each training example, we draw a transition tuple $(o_t, a_{t:t+K-1}, o_{t+K})$ from the dataset $\mathcal{D}$, where the horizon $K$ is sampled from a predefined horizon distribution, instantiated as either uniform or log-uniform in our experiments, and minimize
\begin{equation}
\mathcal{L}(\theta) \;=\; \mathbb{E}\!\left[\, \big\| D\big(g(E(o_t),\, \phi(a_{t:t+K-1}))\big) \,-\, o_{t+K} \big\|^2 \,\right].
\label{eq:k_step_loss}
\end{equation}
A single set of weights is therefore trained to predict observations across the full range of horizons $K \in \{1, \ldots, K_{\max}\}$, rather than separately for any fixed $K$.

This objective stands in contrast to the standard training objective for autoregressive world models, which fits a one-step transition function $\hat{f}_\theta$ against single-step targets:
\begin{equation} 
\mathcal{L}_{\text{1-step}}(\theta) \;=\; \mathbb{E}_{(s_k, a_k, s_{k+1}) \sim \mathcal{D}}\!\left[\, \big\| \hat{f}_\theta(s_k, a_k) - s_{k+1} \big\|^2 \,\right].
\label{eq:1_step_loss}
\end{equation}
Long-horizon predictions are then obtained at inference time by recursively applying $\hat{f}_\theta$ to its own outputs. The two paradigms differ not only in inference behavior, but in two structural ways at training time as well, which we analyze below.

\textbf{Train--inference distribution mismatch.}
The 1-step objective conditions $\hat{f}_\theta$ on ground-truth states $s_k$, but at inference time the model is conditioned on its own previous predictions $\hat{s}_k$ --- a form of exposure bias in which small modeling errors push later inputs progressively off the training distribution. The K-step endpoint loss, by contrast, conditions only on the true initial observation $o_t$ and the true action sequence $a_{t:t+K-1}$ at both training and inference, with no such shift.

\textbf{Misaligned gradient weighting under 1-step training.}
The two objectives also differ in how training gradient is allocated across transitions. Consider what determines endpoint accuracy under autoregressive inference: $\hat{s}_K$ is the $K$-fold composition of $\hat{f}_\theta$ along the action sequence, and its sensitivity to the model parameters expands by the chain rule into:
\begin{equation}
\frac{\partial \hat{s}_K}{\partial \theta} \;=\; \sum_{k=0}^{K-1}\, \underbrace{\Bigg(\prod_{j=k+1}^{K-1} \frac{\partial \hat{f}_\theta(\hat{s}_j, a_j)}{\partial \hat{s}_j}\Bigg)}_{J_{k \to K}}\, \frac{\partial \hat{f}_\theta(\hat{s}_k, a_k)}{\partial \theta},
\label{eq:endpoint_sensitivity}
\end{equation}
where $J_{k \to K}$ is the Jacobian product propagating perturbations~\citep{Pascanu2012OnTD, Bengio94learning} from step $k$ to step $K$. The contribution of each transition to endpoint error is therefore weighted by $J_{k \to K}$, which depends on both the horizon offset $K-k$ and the local Lipschitz behavior of the dynamics. In the expansive regime ($L > 1$), early transitions dominate; in the contractive regime ($L < 1$), late transitions dominate. The 1-step gradient $\nabla_\theta \mathcal{L}_{\text{1-step}}$, however, is a uniform sample average over transitions and contains no Jacobian product. It allocates training signal to each transition without regard for that transition's downstream influence, and is therefore systematically misaligned with the gradient that would actually improve endpoint accuracy. This misalignment grows with $K$: the longer the inference rollout, the more endpoint sensitivity diverges from the uniform 1-step weighting.

The K-step endpoint loss in~\eqref{eq:k_step_loss} avoids this misalignment by definition: its gradient is the derivative of endpoint error with respect to $\theta$, taken in a single backward pass, and structurally matches the sensitivity in~\eqref{eq:endpoint_sensitivity}. We emphasize that this property is a feature of the \emph{training objective}, not of any particular architecture. A learned autoregressive model trained end-to-end on K-step endpoint loss (i.e., backpropagating through a $K$-step unrolled rollout) would inherit the same Jacobian-weighted gradient automatically through the chain rule. Thus, the gains from long-horizon training should be viewed as a property of the direct-prediction paradigm, rather than of a particular architecture. Our architecture provides one practical instantiation that remains tractable to train at large $K$.

\textbf{Why this paradigm is rare in practice.}
Despite these advantages, end-to-end training at long horizons is uncommon. 
For an autoregressive model, computing gradients of an endpoint loss requires backpropagating through \(K\) sequential model applications, leading to increasing memory cost and optimization difficulty as \(K\) grows due to the vanishing and exploding gradient pathologies familiar from recurrent network training.  
Our direct prediction architecture, described next, sidesteps this by making $g$ a non-recursive function of the action sequence, so the gradient path from $\mathcal{L}$ to any model parameter has fixed depth regardless of $K$. This is the role our architectural design plays in the broader story: not as a standalone architectural contribution, but as a practical enabler that makes long-horizon end-to-end training feasible.

\subsection{Model Design}
\label{sec:model_design}

This subsection describes the architecture used to instantiate the four modules of Eq.~\eqref{eq:pipeline}. The design intentionally uses standard components: the observation encoder/decoder, a Transformer encoder~\citep{vaswani2023attention} for the action sequence, and a FiLM-conditioned~\citep{perez2017film} MLP for endpoint dynamics. 

\textbf{Observation encoder/decoder.}
The observation interface is the only modality-specific component of the pipeline. 
For state-based environments, the encoder $E$ is an MLP, and the final output head maps the predicted latent directly back to raw state space. For pixel-based environments, $E$ and $D$ are matched convolutional encoder--decoders.
Both are trained jointly with the action encoder $\phi$ and dynamics module $g$ under the endpoint loss in Eq.~\eqref{eq:k_step_loss}.

\textbf{Action sequence encoder.}
The action encoder $\phi$ maps the full action sequence $a_{t:t+K-1}$ to a fixed-dimensional embedding $z_a \in \mathbb{R}^{D_a}$. Discrete actions are mapped through a learned embedding lookup, and continuous actions through a linear projection. The resulting tokens are processed by a Transformer encoder with rotary positional encodings (RoPE)~\citep{su2023rope}. Since the entire action sequence is available at prediction time, we do not use a causal mask. In our default implementation, the Transformer uses non-causal local self-attention with a window size of $20$, allowing each action token to attend to nearby past and future action tokens within the window. The final-layer action tokens are aggregated by mean pooling to produce the sequence embedding.

\textbf{Dynamics module.}
The dynamics module $g$ predicts the endpoint latent from the encoded initial observation and the action embedding. Let $c=[s_t,z_a]$ denote the concatenated condition. We instantiate $g$ as a stack of FiLM-conditioned residual MLP blocks. An input projection maps $c$ to the hidden dimension; each residual block applies feature-wise affine modulation whose scale and shift parameters are computed from $c$ at every block, re-injecting the conditioning throughout the depth. A final linear head maps the hidden state to the prediction space: raw state space for state-based environments, or latent space before decoding for pixel-based environments.

Importantly, $g$ is non-recursive with respect to the prediction horizon: it predicts the endpoint directly and does not generate any intermediate latent. This keeps the architecture aligned with the endpoint objective in Eq.~\eqref{eq:k_step_loss} and avoids the recursive feedback path used by autoregressive models.

%% file: sections/experiments.tex
\section{Experiments}
\label{sec:experiments}

\begin{figure}
  \centering
  \includegraphics[width=0.9\linewidth]{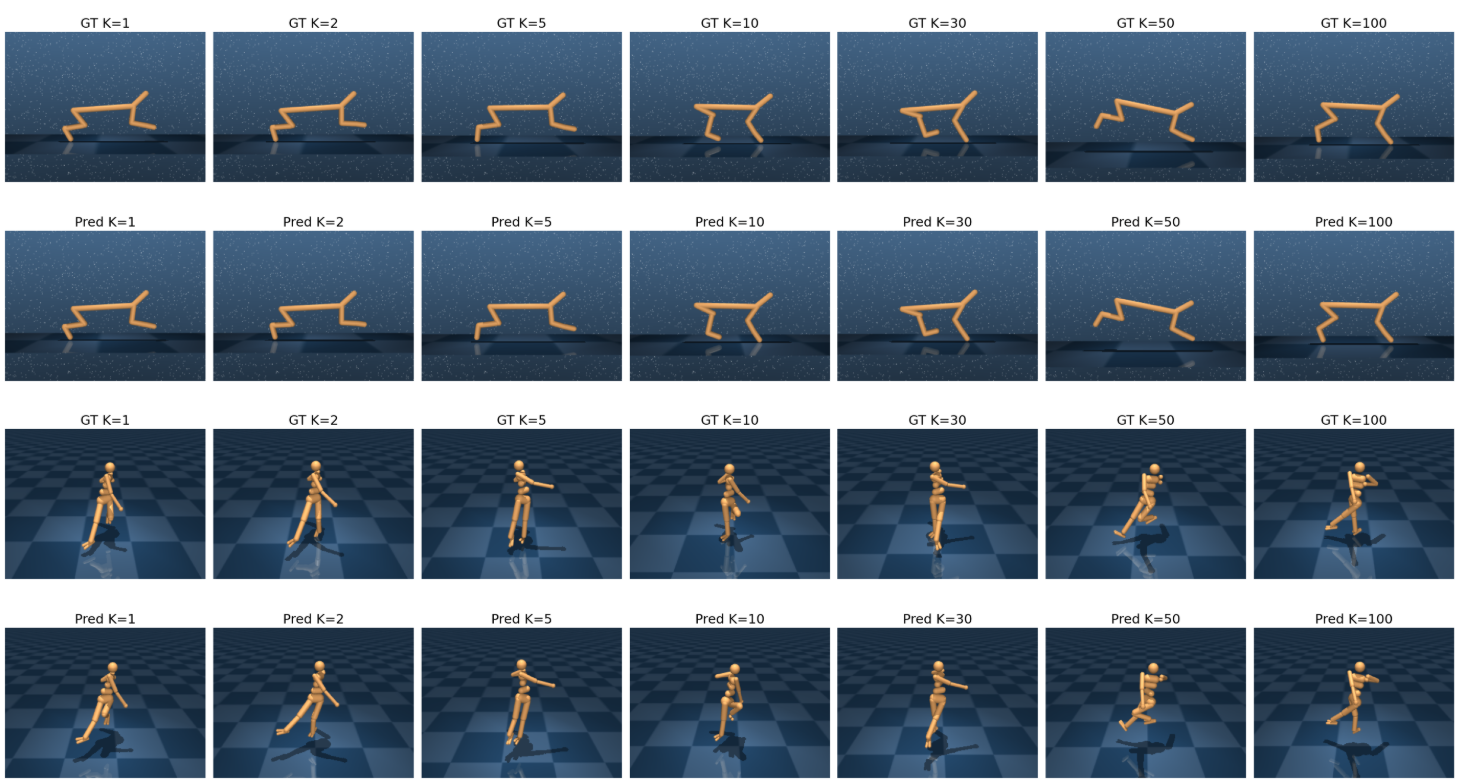}
  \caption{Long-horizon prediction visualizations rendered from states predicted by DPWM$ _{K=100}$, trained and evaluated on the policy dataset.}
\end{figure}

\subsection{Baselines}
\label{sec:baseline}

We compare DPWM with the following representative world-model baselines.

\textbf{MoSim}~\citep{Hao_2025_mosim} is a Neural-ODE-based~\citep{neuralode} motion simulator with physics priors of rigid body dynamics for control tasks. It targets accurate physical state prediction and serves as a strong reference on physics-driven control tasks in the short-horizon regime. 
Because MoSim requires the prediction interval to coincide with the simulator timestep, for the results in Tab.~\ref{tab:prediction-error-main} we set \texttt{action\_repeat}${=}1$ for all methods, rather than the TD-MPC2 defaults. N/A indicates that the MoSim rollout became numerically unstable and did not produce a finite endpoint prediction.

\textbf{ADM}~\citep{admpo, adm2} is the closest any-step dynamics baseline, since it also predicts the endpoint state at a variable horizon. However, it differs from our approach in two key respects. First, it recurrently processes the action sequence through latent intermediate states. Second, in its published setting, it is trained with very short prediction horizons, e.g., $K=2$ or $3$ on the Cheetah and Humanoid tasks. In contrast, our focus is end-to-end supervision at much longer horizons. To disentangle these two factors, we evaluate ADM in two settings. 
The main results below reproduce its published short-horizon training; 
Sec.~\ref{sec:state_control} additionally reports ADM retrained under our 
long-horizon objective at $K_{\max}\!=\!100$, providing a controlled 
comparison between the training paradigm and architecture.

For one-step prediction models 
we evaluate by autoregressive rollout: each predicted observation is fed back into the model with the next action in the sequence to produce the following prediction, repeated for $K$ steps to reach the endpoint.
For ADM, we obtain the predicted endpoint using the rollout generation procedure from the original paper.
Together, these baselines allow us to test whether long-horizon endpoint accuracy is better addressed by direct endpoint supervision than by short-horizon supervision followed by autoregressive rollout.

\begin{table}[t]
\caption{Raw state-space MSE of endpoint observation prediction.}
\label{tab:prediction-error-main}
\centering
\small
\setlength{\tabcolsep}{5pt}
\resizebox{0.8\textwidth}{!}{%
\begin{tabular}{llccc ccc}
\toprule
& & \multicolumn{3}{c}{rand train, rand eval}
  & \multicolumn{3}{c}{policy train, policy eval} \\
\cmidrule(lr){3-5}
\cmidrule(lr){6-8}
Env & Horizon
& DPWM$_{K=100}$ & ADM$_{K=3}$ & MoSim
& DPWM$_{K=100}$ & ADM$_{K=3}$ & MoSim \\
\midrule

\multirow{4}{*}{Cheetah}
& 1   & \textbf{0.0277} & 0.0345 & 0.0519
      & \textbf{0.1888} & 0.2624 & 0.2801 \\
& 16  & \textbf{0.0920} & 1.8550 & 0.2331
      & \textbf{0.2918} & 17.3993 & 1.8238 \\
& 100 & \textbf{0.1854} & 1.7240 & 0.3215
      & \textbf{0.4452} & 25.6086 & 11.0004 \\
& 200 & 0.5110 & 1.7875 & \textbf{0.3190}
      & \textbf{1.3840} & 34.0766 & N/A \\
\midrule

\multirow{4}{*}{Humanoid}
& 1   & \textbf{0.2449} & 0.2724 & 0.3378
      & \textbf{0.2812} & 0.3910 & 0.6887 \\
& 16  & \textbf{0.9172} & 10.8206 & 3.0627
      & \textbf{1.9022} & 28.5179 & 9.2534 \\
& 100 & \textbf{2.6677} & 17.9088 & 4.5153
      & \textbf{5.7811} & 49.4643 & 17.4877 \\
& 200 & \textbf{3.2823} & 18.2562 & 5.2478
      & \textbf{7.7810} & 51.0466 & 20.4348 \\
\midrule

\multirow{4}{*}{Hopper}
& 1   & 0.3727 & 0.4960 & \textbf{0.0105}
      & 0.0820 & \textbf{0.0253} & 0.0452 \\
& 16  & 0.4050 & 0.6727 & \textbf{0.1887}
      & \textbf{0.1249} & 5.4239 & 0.5634 \\
& 100 & \textbf{0.4081} & 1.0853 & 0.4158
      & \textbf{0.2733} & 14.9331 & 30.2598 \\
& 200 & \textbf{0.5975} & 1.1014 & 0.5982
      & \textbf{2.7208} & 15.3578 & N/A \\
\midrule

\multirow{4}{*}{Walker}
& 1   & \textbf{0.0622} & 0.0780 & 0.0686
      & 0.1907 & \textbf{0.1180} & 0.3475 \\
& 16  & \textbf{1.7475} & 2.4299 & 1.7650
      & \textbf{2.0393} & 14.9711 & 11.9352 \\
& 100 & \textbf{1.1604} & 3.8802 & 15.3791
      & \textbf{3.7131} & 55.7769 & 33.7685 \\
& 200 & \textbf{1.6793} & 3.8563 & N/A
      & \textbf{18.9293} & 44.1142 & N/A \\
\bottomrule
\end{tabular}
}
\end{table}

\subsection{State-based Continuous Control}
\label{sec:state_control}

\textbf{Setup.}
We evaluate on four tasks from the DeepMind Control Suite~\citep{tunyasuvunakool2020}: \texttt{cheetah-run}, \texttt{humanoid-walk}, \texttt{hopper-hop}, and \texttt{walker-run}. For each task, the raw observation consists of the physics state $(q_{\text{pos}}, q_{\text{vel}})$. The corresponding dimensions of $(q_{\text{pos}}, q_{\text{vel}})$ are $(9, 9)$, $(28, 27)$, $(7, 7)$, and $(9, 9)$, respectively, and the action spaces are $\mathbb{R}^{6}$, $\mathbb{R}^{21}$, $\mathbb{R}^{4}$, and $\mathbb{R}^{6}$.

For each environment we construct training and evaluation datasets distinguished by the action distribution used to roll out trajectories. The \emph{random} dataset draws actions uniformly from the action space. The \emph{policy} dataset uses a publicly released TD-MPC2~\citep{hansen2024tdmpc2} expert checkpoint. A pure expert, however, produces highly consistent trajectories that converge to a narrow distribution of states and actions, particularly over long horizons. We mitigate this with an \emph{intermix} protocol that injects short random segments into expert rollouts, enriching the diversity of trajectories seen during training.

We evaluate prediction accuracy under two matched training and evaluation action distributions: random-policy training and evaluation, and intermix-policy training and evaluation . The main evaluation metric is the \emph{endpoint} mean squared error between the predicted observation $\hat{o}_{t+K}$ and the ground-truth observation $o_{t+K}$. We report endpoint MSE rather than averaging across the rollout because compounding errors in autoregressive inference (Sec.~\ref{sec:formulation}) are most pronounced at the horizon endpoint; averaging across intermediate steps would dilute this signal with easier short-horizon predictions. We evaluate at horizons $K \in \{1, 16, 100, 200\}$.

DPWM is trained with our long-horizon end-to-end endpoint objective using $K_{\max}=100$. ADM is trained using its published short-horizon setting with $K_{\max}=3$. Autoregressive baselines are trained with local one-step prediction losses, corresponding to $K_{\max}=1$, and are evaluated by rolling out their predictions to the target horizon.

\textbf{Evaluation results.} 
DPWM achieves substantially lower endpoint MSE as the prediction horizon increases across all four tasks (Tab.~\ref{tab:prediction-error-main}), while ADM and MoSim are competitive mainly at short horizons. 
This trend reflects a fundamental difference in how long-range prediction is learned. 
ADM composes short transitions recursively, so small prediction errors alter subsequent inputs and are progressively amplified over the rollout. 
MoSim introduces an explicit physics prior that can improve local prediction, but systematic model mismatch in the learned physical dynamics can still accumulate over long horizons and eventually cause divergence. 
In contrast, DPWM directly learns the effect of an entire action sequence under multi-horizon endpoint supervision, aligning its training objective with long-horizon evaluation while avoiding repeated self-conditioning on predicted states. 
These results suggest that accurate long-horizon prediction requires explicitly learning the long-range transition, rather than relying solely on short-step accuracy or physics-informed priors.

\begin{figure}
  \centering
  \includegraphics[width=0.85\linewidth]{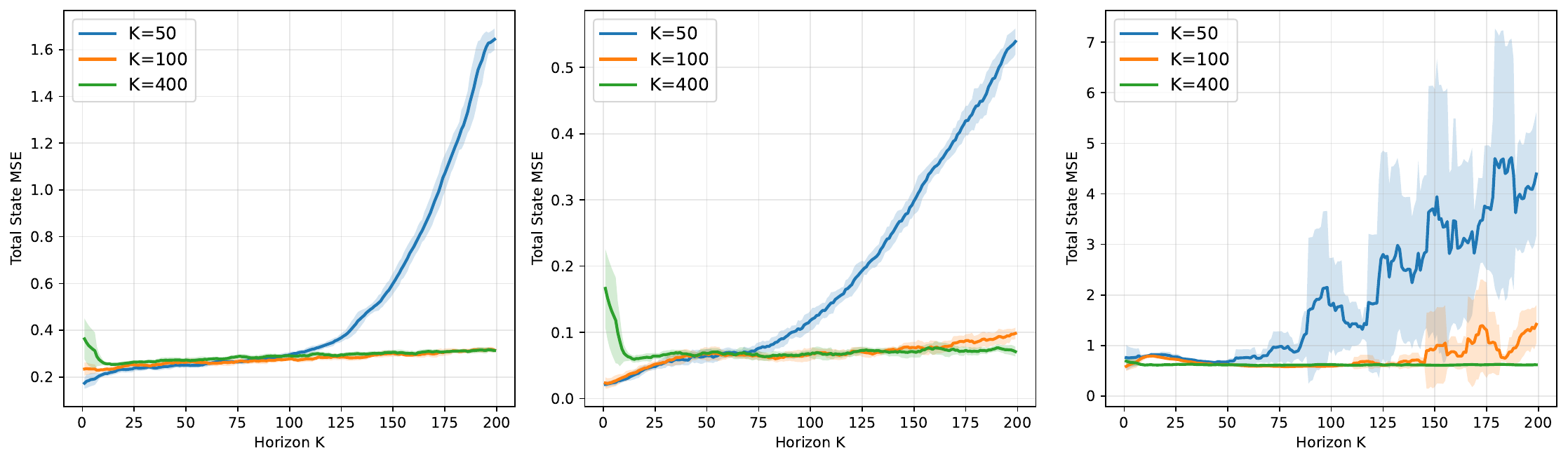}
  \caption{Endpoint prediction MSE computed in \emph{normalized} state space across evaluation horizons for models trained with $K_{\max}\in\{50,100,400\}$ on intermix-policy data. 
Results are shown for \texttt{humanoid\_walk} evaluated on intermix-policy (left), \texttt{cheetah\_run} on intermix-policy (middle), and \texttt{cheetah\_run} on random-policy (right).
}
\label{fig:kselection}
\end{figure}

\textbf{Selection of training parameter $K_{\max}$.}
We compare DPWM trained with $K_{\max} \in \{50, 100, 400\}$ on intermix-policy data, evaluated across $K \in [1, 200]$ under three settings: in-distribution evaluation on \texttt{humanoid\_walk} and \texttt{cheetah\_run}, and out-of-distribution evaluation on \texttt{cheetah\_run} with random-action trajectories (Fig.~\ref{fig:kselection}). 
Models trained with $K_{\max}\!=\!50$ perform well only within their training horizon and exhibit sharp error growth beyond it, suggesting that long-horizon prediction requires long-horizon training. Models trained with $K_{\max}\!=\!100$ remain stable across the full evaluation range, including the $K\!=\!200$ extrapolation regime. Models trained with $K_{\max}\!=\!400$ are comparable to $K_{\max}\!=\!100$ at most in-distribution horizons but exhibit elevated error at very short horizons; under out-of-distribution evaluation on random-action trajectories, however, $K_{\max}\!=\!400$ yields consistently lower error than $K_{\max}\!=\!100$, providing additional evidence that longer training horizons yield stronger generalization to unseen action distributions. We adopt $K_{\max}\!=\!100$ as the default for all main experiments based on its balanced performance across horizons.

\textbf{Training objective paradigm dominates architecture choice.}
To isolate the contribution of our long-horizon training paradigm from the 
architectural choice (Transformer vs.\ RNN), we compare three models on 
\texttt{humanoid\_walk} in Tab.~\ref{tab:adm}: DPWM$ _{K=100}$, ADM$ _{K=100}$ that aligns 
with DPWM setting, and ADM$ _{K=3}$ that aligns with the published setting. 
Retraining ADM under our long-horizon paradigm substantially closes its 
long-range prediction gap to DPWM at evaluation horizons $K\!\ge\!16$. ADM$ _{K=3}$
and ADM$ _{K=100}$ share the same encoder and inference procedure, differing only in 
the training horizon. Short-horizon supervision propagates gradient through 
only a handful of dynamics steps and offers no training pressure on structure 
that emerges over longer horizons. Long-horizon end-to-end supervision, by 
attaching the loss directly to a $K$-step prediction, forces the encoder to 
represent features whose evolution remains accurate at horizons that local 
losses never reach. 

\begin{table}[t]
  \caption{Endpoint prediction MSE on \texttt{humanoid\_walk}, computed in \emph{normalized} state space. The \texttt{action\_repeat} follows the TD-MPC2 default setting.
  The suffix indicates the training horizon $K_{\max}$, with ADM$_{K=3}$ following the published ADM setting.}
  \label{tab:adm}
  \centering
  \small
  \setlength{\tabcolsep}{5pt}
  \begin{tabular}{llccc ccc}
    \toprule
    & & \multicolumn{3}{c}{Policy train, policy eval}
      & \multicolumn{3}{c}{Policy train, rand eval} \\
    \cmidrule(lr){3-5}
    \cmidrule(lr){6-8}
    Env & Horizon
    & DPWM$ _{K=100}$ & ADM$ _{K=100}$ & ADM$ _{K=3}$
    & DPWM$ _{K=100}$ & ADM$ _{K=100}$ & ADM$ _{K=3}$ \\
    \midrule
    \multirow{4}{*}{Humanoid}
    & 1   & 0.2700 & 0.3448 & \textbf{0.1385}
          & 0.4560 & 0.5639 & \textbf{0.2637} \\
    & 16  & \textbf{0.2394} & 0.2577 & 0.4318
          & \textbf{0.4340} & 0.4732 & 0.5763 \\
    & 100 & \textbf{0.2658} & 0.2999 & 0.7756
          & \textbf{0.6058} & 0.6318 & 0.8011 \\
    & 200 & \textbf{0.3234} & 0.3483 & 0.8882
          & \textbf{0.6305} & 0.6622 & 0.9701 \\
    \bottomrule
  \end{tabular}
\end{table}

\begin{figure}
  \centering
  \includegraphics[width=\linewidth]{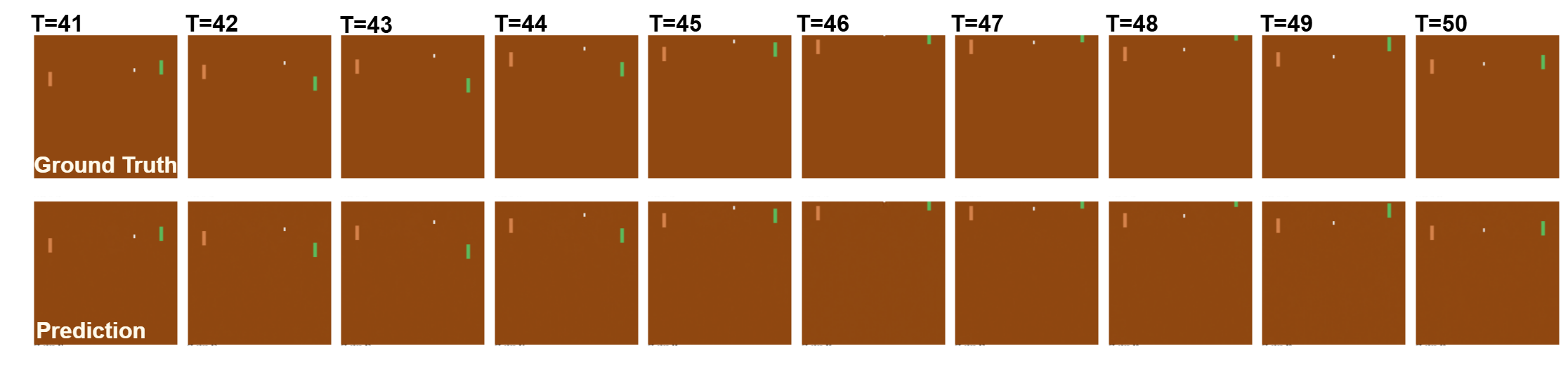}
  \caption{DPWM trajectory generation on Pong. Top: ground-truth observations. 
  Bottom: per-step predictions $\hat{o}_t$, each independently generated from 
  the same initial observation $o_0$ and the action prefix $a_{0:t-1}$. 
  Frames $T=41$ to $T=50$ from a $70$-step rollout; full rollout in 
  App.~\ref{app:viz}.}
\label{fig:pong}
\end{figure}

\subsection{Pixel-Based Discrete Control}
\label{sec:atari}

\textbf{Setup.}
We evaluate on \texttt{Pong} from the Atari Learning Environment~\citep{Bellemare_2013}. Observations are single-frame $160 \times 160$ RGB images cropped from the native $210 \times 160$ to remove the static scoreboard region; we apply an action repeat of $4$ but no frame stacking and no auxiliary state information, so each prediction is conditioned on a single observed frame. The action space contains $4$ discrete actions.

For training data, we use a deterministic chase-ball heuristic policy: the green paddle tracks the ball's vertical position, except when the ball lies within a narrow band, where the paddle stays still. This deliberate miss zone produces approximately tied scores against the built-in opponent, balancing the distribution of game states in the training data. DPWM is trained with $K_{\max}=100$ on this policy data and evaluated on held-out trajectories from the same distribution. 

\textbf{Results.}
Fig.~\ref{fig:pong} shows a 10-frame slice of a 70-step trajectory generated by DPWM$ _{K=100}$; the rollout terminates around $T=70$ where the held-out episode ends and the game state resets. Although DPWM is trained to predict only the endpoint at horizon $K$, the same model can generate a per-step trajectory by feeding successively longer slices of the action sequence (the first $t$ actions, for $t=1, \ldots, K$) into the model, each query conditioned on the same initial observation $o_0$. 
Each frame is produced by an independent forward pass rather than by recursively feeding previous predictions back into the model, so the trajectory visualization does not rely on autoregressive rollout.

Across the sequence, the predicted frames remain visually close to the ground truth. This suggests that endpoint-supervised direct prediction can be queried at multiple horizons to provide intermediate states, making it compatible with downstream pipelines that require trajectory-level predictions, such as model-based planning.

%% file: sections/conclusion.tex
\section{Related Work}

\textbf{World Models.}
A world model is an action-conditioned generative model of environment dynamics: given an initial state and a sequence of actions, it predicts the resulting future observations~\citep{ha2018world, sutton1991dyna, lecun2022path, bruce2024genie}. Such models serve as learned surrogates for the environment, supporting downstream uses including planning~\citep{hansen2022tdmpc, hansen2024tdmpc2} and model-based reinforcement learning~\citep{hafner2019dreamerv1, hafner2020dreamerv2, hafner2025dreamerv3, sutton1991dyna, henaff2017model, micheli2023iris, alonso2024diamond}. 
Recent work targets multiple dimensions of model quality including physical fidelity, prediction accuracy, temporal coherence over rollouts, and generalization to new states and action distributions~\citep{kang2025farvideogenerationworld, Liu2026FromKT,li2025smallworld, xia2026cloningdeterministicworldscritical, Hao_2025_mosim}.
Across these settings, however, many widely used systems still rely on one-step or short-horizon transition prediction, with long-horizon behavior obtained at inference time via recursive rollout.

\textbf{Multi-Step Prediction and Action Sequence Modeling.}
A line of work~\citep{eysenbach2018diversity, machado2023temporalabstractionreinforcementlearning} compresses consecutive low-level actions into higher-level units such as learned skills, and plans or predicts at this level rather than at the primitive-action level. Another direction processes raw action sequences to reduce the number of autoregressive prediction steps without explicit semantic abstraction: some~\citep{zhang2023leveragingjumpymodelsplanning, zhao2023learning} predict states at a fixed horizon offset, while variable-length self-segmenting models~\citep{gumbsch2024learning, jayaraman2018timeagnosticpredictionpredictingpredictable} learn to find the next endpoint in an unsupervised manner. 
Hierarchical latent world models~\citep{zhang2026hierarchicalplanninglatentworld} also encode variable-length action sequences with a Transformer encoder, but they operate at shorter horizons and focus on downstream task performance. 
These works either focus on finding good segmentation points or operate at short horizons; DPWM instead predicts over variable-length action sequences without requiring a temporal segmentation prior, training at extended horizons and processing the entire sequence non-recursively in a single forward pass.

\section{Conclusion and Limitations}

We examined long-horizon endpoint supervision as an alternative to local transition training for world models. To make this paradigm practical, we proposed DPWM, a non-recursive model that directly predicts the endpoint observation from an initial observation and an action sequence. 
Across continuous-control and pixel-based benchmarks, DPWM improves endpoint accuracy at extended horizons. 
Our ablations further show that the benefit is not tied to the DPWM backbone alone: other model designs, including recurrent ones, also improve under the same end-to-end endpoint objective. 
These results point to supervision horizon as an important design choice for world models used in long-range prediction.

\textbf{Limitations and future directions.} Our evaluation focuses on endpoint prediction rather than downstream control performance, and independently queried predictions are not explicitly constrained to form temporally consistent trajectories. Future work could evaluate DPWM in planning and policy learning, introduce trajectory-level consistency constraints, and extend the framework to stochastic environments.

%% file: sections/appendix.tex
\newpage
\appendix

\section{Lipschitz Proof of Error Bound}
\label{app:lipschitz}

We now compare how prediction errors scale with the horizon $K$ under the two
modeling paradigms. Throughout this analysis, we assume the dynamics $f$ are
$L$-Lipschitz in the state argument: for all $s, s' \in \mathcal{S}$ and
$a \in \mathcal{A}$,
\begin{equation}
    \|f(s, a) - f(s', a)\| \leq L \, \|s - s'\|.
\end{equation}

\textbf{Recursive rollout.}
Let $\hat{f}$ denote a learned one-step model with bounded approximation error
$\|\hat{f}(s, a) - f(s, a)\| \leq \epsilon$ for all $(s, a)$. Define the
$k$-step rollout error as $e_k := \|\hat{s}_{t+k} - s_{t+k}\|$, where
$\hat{s}_{t+k}$ is obtained by recursively applying $\hat{f}$ to its own
predictions. A standard triangle inequality yields
\begin{equation}
\begin{aligned}
    e_{k+1}
    &= \|\hat{f}(\hat{s}_{t+k}, a_{t+k}) - f(s_{t+k}, a_{t+k})\| \\
    &\leq \underbrace{\|\hat{f}(\hat{s}_{t+k}, a_{t+k}) - f(\hat{s}_{t+k}, a_{t+k})\|}_{\leq\,\epsilon}
    + \underbrace{\|f(\hat{s}_{t+k}, a_{t+k}) - f(s_{t+k}, a_{t+k})\|}_{\leq\,L\,e_k} \leq \epsilon + L \, e_k.
\end{aligned}
\end{equation}
Unrolling the recursion from $e_0 = 0$ gives
\begin{equation}
    e_K \;\leq\; \epsilon \sum_{i=0}^{K-1} L^i
    \;=\;
    \begin{cases}
        \epsilon \, \dfrac{L^K - 1}{L - 1}, & L \neq 1, \\[4pt]
        K\,\epsilon, & L = 1.
    \end{cases}
    \label{eq:rollout_bound}
\end{equation}

This compounding-error bound has been observed in various forms across the
model-based RL literature, which is linear in $K$ when $L = 1$, and grows exponentially in $K$ when $L > 1$. Crucially, this amplification arises not from the modeling error
$\epsilon$ itself, but from the recursive structure of rollout: each step
re-applies $f$ to a perturbed input, and the Lipschitz constant $L$ propagates
the perturbation forward. For control tasks with locally expansive dynamics (e.g., chaotic systems, contact-rich manipulation), $L > 1$ is the norm rather than the exception, making this error amplification a \emph{structural} obstacle to long-horizon prediction.

\textbf{Direct prediction.}
Let $g_\theta : \mathcal{S} \times \mathcal{A}^K \to \mathcal{S}$ denote our
direct prediction model, which approximates $f^{(K)}$ as a single mapping.
Denote its endpoint approximation error by
\begin{equation}
    \delta_K \;:=\;
    \sup_{s_t,\, a_{t:t+K-1}}
    \big\| g_\theta(s_t, a_{t:t+K-1}) - f^{(K)}(s_t, a_{t:t+K-1}) \big\|.
\end{equation}
The endpoint error of direct prediction is then bounded by $\delta_K$ alone,
with no recursive amplification term. Comparing with~\eqref{eq:rollout_bound},
direct prediction yields a tighter long-horizon bound whenever
\begin{equation}
    \delta_K \;<\; \epsilon \sum_{i=0}^{K-1} L^i.
    \label{eq:sufficient_condition}
\end{equation}

We emphasize that direct prediction does not eliminate the difficulty of long-horizon modeling; it relocates it. Recursive rollout incurs a horizon-dependent \emph{inference-time} amplification through the recursive factor $\sum_{i=0}^{K-1}L^i$ in~\eqref{eq:rollout_bound}. Direct prediction avoids this recursive amplification, but pays a \emph{modeling} cost: as $K$ grows, the target map $f^{(K)}$ can become harder to approximate, and the endpoint error $\delta_K$ may grow with the prediction horizon. The substantive question is therefore empirical: \emph{how fast does $\delta_K$ grow?} When $L>1$, if $\delta_K$ grows more slowly than the rollout bound $\epsilon\sum_{i=0}^{K-1}L^i$, direct prediction can become favorable at long horizons. When $L=1$, the corresponding comparison is against the linear bound $K\epsilon$. For contractive dynamics with $L<1$, the rollout bound is less severe, and the advantage of direct prediction must be established empirically.

\section{DPWM Prediction Visualization}
\label{app:viz}

\begin{figure}
  \centering
  \includegraphics[width=\linewidth]{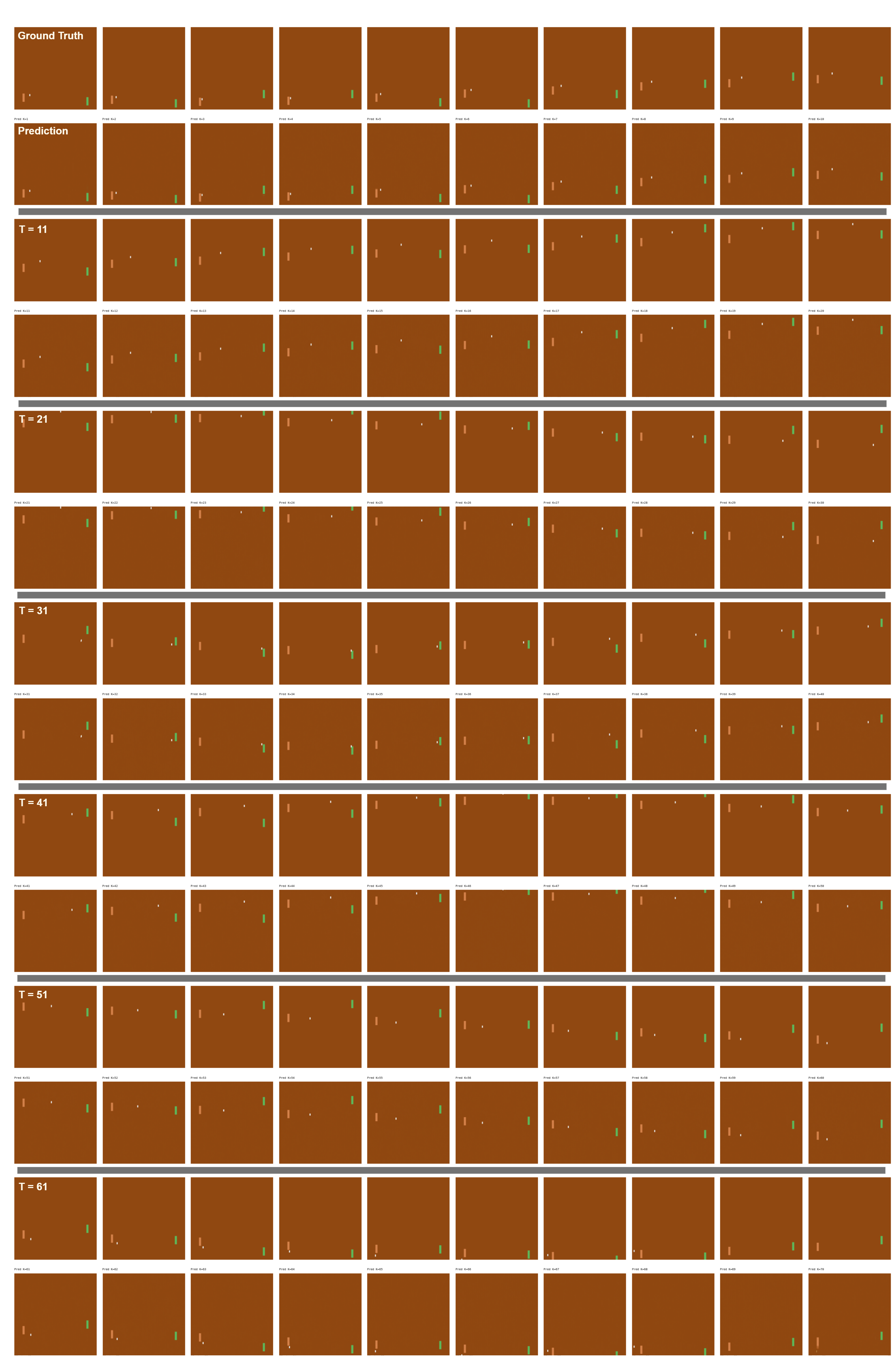}
  \caption{Full DPWM 70-step trajectory generation on Pong. Top: ground-truth observations. 
  Bottom: per-step predictions $\hat{o}_t$, each independently generated from 
  the same initial observation $o_0$ and the action prefix $a_{0:t-1}$. }
\label{fig:pong-ori}
\end{figure}

Fig.~\ref{fig:pong-ori} shows the full 70-step Pong trajectory corresponding to the 10-frame slice in Fig.~\ref{fig:pong}. DPWM is trained on trajectories from a simple chase-ball policy and evaluated here on a held-out trajectory from the same policy. The sequence is obtained by querying the model at successive horizons, where the prediction at time $t$ is conditioned on $o_0$ and $a_{0:t-1}$ rather than on previously predicted frames.

\section{Hyperparameter and Implementation Details}
\label{app:hyper}

\begin{table}[t]
\centering
\caption{Default DPWM hyperparameters and implementation settings.}
\label{tab:hparams}
\begin{tabular}{lll}
\toprule
\textbf{Component} & \textbf{Hyperparameter} & \textbf{Value} \\
\midrule
\multirow{5}{*}{State observation}
& Cheetah state/action dim      & $18$ / $6$ \\
& Humanoid state/action dim     & $55$ / $21$ \\
& Hopper state/action dim      & $14$ / $4$ \\
& Walker state/action dim     & $18$ / $6$ \\
& Encoder                       & 4-layer MLP, width $256$ \\
\midrule
\multirow{5}{*}{Pixel observation}
& Resolution (crop)             & $160\times160$ (from $210\times160$) \\
& Action repeat / discrete acts & $4$ / $4$ \\
& Conv/deconv blocks            & $5$ / $5$ \\
& Channel widths                & $32,64,128,256,256$ \\
& Latent dim                    & $256$ \\
\midrule
\multirow{6}{*}{Transformer Action encoder}
& Token dim $D_a$               & $256$ \\
& Layers                        & $5$\\
& Heads                         & $4$ \\
& Head dim                      & $64$ \\
& FFN expansion / dropout       & $4\times$ / $0.1$ \\
& Attention window size         & $20$ \\
\midrule
\multirow{3}{*}{Dynamics module}
& Condition                     & $[s_t,z_a]$ \\
& Hidden width            & $512$ \\
& Hidden blocks           & $10$ \\
\midrule
\multirow{8}{*}{Optimization}
& Optimizer                     & Adam \\
& Learning rate                 & $3\times10^{-4}$ \\
& LR schedule                   & cosine to $10^{-10}$ \\
& State batch size              & $256$ \\
& Pixel batch size              & $128$ \\
& Gradient clipping             & $1.0$ \\
& Epochs                        & $3600$ \\
& Max updates per epoch         & $2000$ state / $500$ pixel \\
\bottomrule
\end{tabular}
\end{table}

Table~\ref{tab:hparams} summarizes the architecture and training configuration of DPWM. All modules are
trained jointly with the endpoint reconstruction loss mentioned in Sec.~\ref{sec:training}.

\paragraph{Observation encoder and decoder.}
For state-based DeepMind Control experiments, the observation encoder is a 4-layer MLP with
width 256 and ReLU activations, mapping the raw physics state $o_t$ to a latent representation
$s_t\in\mathbb{R}^{256}$. For these experiments, the
dynamics output head maps directly back to raw state space, where the endpoint MSE is computed.

For pixel-based Pong experiments, we use a matched convolutional encoder--decoder. The observation
is a single RGB frame cropped from the native $210\times160$ resolution to $160\times160$ to remove
the static scoreboard region, with an action repeat of 4 and no frame stacking. Pixel values are
normalized to $[0,1]$, and the reconstruction loss is computed in this normalized space (the decoder
ends in a sigmoid). The encoder applies five stride-2 convolutional blocks with kernel size 4,
padding 1, GroupNorm, and SiLU activations, using channel widths $32,64,128,256,256$. The resulting
$5\times5$ feature map is flattened and projected to a 256-dimensional latent. 

\paragraph{Action sequence encoder.}
The default action encoder is a RoPE Transformer. Continuous actions are projected linearly to action
tokens, while discrete actions are mapped through a learned embedding lookup. The token dimension
is $D_a=256$. The Transformer uses 5 layers, 4 attention heads, head dimension 64, a $4\times$
feed-forward expansion with GELU activations, and dropout 0.1. We use rotary positional encodings (RoPE) and bidirectional (non-causal)
local self-attention with window size 20. The final action embedding
$z_a\in\mathbb{R}^{256}$ is obtained by mean-pooling the final-layer action tokens.

\paragraph{Dynamics module.}
The dynamics module is a FiLM-conditioned residual MLP. Given the encoded observation $s_t$ and
the action embedding $z_a$, we form the condition $c=[s_t,z_a]$. The condition is projected to
the hidden width with a linear layer and SiLU activation, then processed by a stack of FiLM residual
blocks. Each block applies LayerNorm, a linear layer, SiLU activation, feature-wise affine modulation,
and a residual connection. We use hidden width 512 and 10 blocks. 

\paragraph{Training.}
We train with Adam using learning rate $3\times10^{-4}$ and gradient clipping with maximum norm 1.0. The learning rate follows cosine annealing to $10^{-10}$ over 3600 epochs. State-based tasks use batch size 256, and pixel-based tasks use 128.

\section{Quantitative Pong Evaluation}
\label{sec:pong-quantitative}

The main paper provides only qualitative Pong rollouts (Fig.~\ref{fig:pong}). Here we report quantitative pixel-space MSE at multiple horizons, comparing $\text{DPWM}_{K=100}$ against $\text{ADM}_{K=100}$ as the architecture-controlled long-horizon baseline.

For every target horizon $t$, we query each model with the same initial frame $o_0$ and the action prefix $a_{0:t-1}$. For DPWM, each entry is produced by an independent forward pass—no autoregressive rollout, no intermediate latents. Pixel MSE is computed on the cropped $160 \times 160$ RGB frame.

\begin{table}[h]
\centering
\caption{Pong endpoint pixel MSE ($\times 10^{-4}$).}
\label{tab:pong-pixel-mse}
\begin{tabular}{ccc}
\toprule
Horizon & $\text{DPWM}_{K=100}$ & $\text{ADM}_{K=100}$ \\
\midrule
1   & $1.1759$ & $\mathbf{1.1252}$ \\
16  & $\mathbf{5.5467}$ & $5.6842$ \\
100 & $\mathbf{5.9387}$ & $6.8619$ \\
200 & $\mathbf{5.5852}$ & $6.9098$ \\
400 & $\mathbf{5.9844}$ & $6.7410$ \\
\bottomrule
\end{tabular}
\end{table}

%% file: main.bib
@String(CVPR= {IEEE Conf. Comput. Vis. Pattern Recog.})

@String(NIPS= {Adv. Neural Inform. Process. Syst.})

@String(ICLR = {Int. Conf. Learn. Represent.})

@String(CVPR  = {CVPR})

@String(NIPS  = {NeurIPS})

@String(ICLR  = {ICLR})

@inproceedings{Pascanu2012OnTD,
  title={On the difficulty of training recurrent neural networks},
  author={Razvan Pascanu and Tomas Mikolov and Yoshua Bengio},
  booktitle={International Conference on Machine Learning},
  year={2012},
  url={https://api.semanticscholar.org/CorpusID:14650762}
}

@misc{
benechehab2024multitimestep,
title={Multi-timestep models for Model-based Reinforcement Learning},
author={Abdelhakim Benechehab and Giuseppe Paolo and Albert Thomas and Maurizio Filippone and Bal{\'a}zs K{\'e}gl},
year={2024},
url={https://openreview.net/forum?id=Rh4DmXaf8R}
}

@inproceedings{lamb2016professorforcing,
author = {Goyal, Anirudh and Lamb, Alex and Zhang, Ying and Zhang, Saizheng and Courville, Aaron and Bengio, Yoshua},
title = {Professor forcing: a new algorithm for training recurrent networks},
year = {2016},
isbn = {9781510838819},
publisher = {Curran Associates Inc.},
address = {Red Hook, NY, USA},
booktitle = {Proceedings of the 30th International Conference on Neural Information Processing Systems},
pages = {4608–4616},
numpages = {9},
location = {Barcelona, Spain},
series = {NIPS'16}
}

@article{Vafa2025WhatHA,
  title={What Has a Foundation Model Found? Using Inductive Bias to Probe for World Models},
  author={Keyon Vafa and Peter G. Chang and Ashesh Rambachan and Sendhil Mullainathan},
  journal={ArXiv},
  year={2025},
  volume={abs/2507.06952},
  url={https://api.semanticscholar.org/CorpusID:280150828}
}

@article{Liu2026FromKT,
  title={From Kepler to Newton: Inductive Biases Guide Learned World Models in Transformers},
  author={Ziming Liu and Sophia Sanborn and Surya Ganguli and Andreas Tolias},
  journal={ArXiv},
  year={2026},
  volume={abs/2602.06923},
  url={https://api.semanticscholar.org/CorpusID:285401944}
}

@article{Finn2016UnsupervisedLF,
  title={Unsupervised Learning for Physical Interaction through Video Prediction},
  author={Chelsea Finn and Ian J. Goodfellow and Sergey Levine},
  journal={ArXiv},
  year={2016},
  volume={abs/1605.07157},
  url={https://api.semanticscholar.org/CorpusID:2659157}
}

@inproceedings{micheli2023iris,
  title = {Transformers are Sample-Efficient World Models},
  author = {Micheli, Vincent and Alonso, Eloi and Fleuret, Fran{\c{c}}ois},
  booktitle = {International Conference on Learning Representations},
  year = {2023}
}

@inproceedings{alonso2024diamond,
  title = {{DIAMOND}: Diffusion for World Modeling: Visual Details Matter in Atari},
  author = {Alonso, Eloi and Jelley, Adam and Micheli, Vincent and Kanervisto, Anssi and Storkey, Amos and Pearce, Tim and Fleuret, Fran{\c{c}}ois},
  booktitle = {Advances in Neural Information Processing Systems},
  year = {2024}
}

@inproceedings{hafner2019planet,
  title = {Learning Latent Dynamics for Planning from Pixels},
  author = {Hafner, Danijar and Lillicrap, Timothy and Fischer, Ian and Villegas, Ruben and Ha, David and Lee, Honglak and Davidson, James},
  booktitle = {Proceedings of the 36th International Conference on Machine Learning},
  pages = {2555--2565},
  year = {2019},
  volume = {97},
  series = {Proceedings of Machine Learning Research}
}

@misc{li2025smallworld,
      title={SmallWorlds: Assessing Dynamics Understanding of World Models in Isolated Environments}, 
      author={Xinyi Li and Zaishuo Xia and Weyl Lu and Chenjie Hao and Yubei Chen},
      year={2025},
      eprint={2511.23465},
      archivePrefix={arXiv},
      primaryClass={cs.LG},
      url={https://arxiv.org/abs/2511.23465}, 
}

@misc{xia2026cloningdeterministicworldscritical,
      title={Cloning Deterministic Worlds: The Critical Role of Latent Geometry in Long-Horizon World Models}, 
      author={Zaishuo Xia and Yukuan Lu and Xinyi Li and Yifan Xu and Yubei Chen},
      year={2026},
      eprint={2510.26782},
      archivePrefix={arXiv},
      primaryClass={cs.LG},
      url={https://arxiv.org/abs/2510.26782}, 
}

@InProceedings{Hao_2025_mosim,
    author    = {Hao, Chenjie and Lu, Weyl and Xu, Yifan and Chen, Yubei},
    title     = {Neural Motion Simulator Pushing the Limit of World Models in Reinforcement Learning},
    booktitle = {Proceedings of the Computer Vision and Pattern Recognition Conference (CVPR)},
    month     = {June},
    year      = {2025},
    pages     = {27608-27617}
}

@article{ha2018world,
  title={World models},
  author={Ha, David and Schmidhuber, J{\"u}rgen},
  journal={arXiv preprint arXiv:1803.10122},
  year={2018}
}

@misc{hansen2024tdmpc2,
      title={TD-MPC2: Scalable, Robust World Models for Continuous Control}, 
      author={Nicklas Hansen and Hao Su and Xiaolong Wang},
      year={2024},
      eprint={2310.16828},
      archivePrefix={arXiv},
      primaryClass={cs.LG},
      url={https://arxiv.org/abs/2310.16828}, 
}

@misc{hansen2022tdmpc,
      title={Temporal Difference Learning for Model Predictive Control}, 
      author={Nicklas Hansen and Xiaolong Wang and Hao Su},
      year={2022},
      eprint={2203.04955},
      archivePrefix={arXiv},
      primaryClass={cs.LG},
      url={https://arxiv.org/abs/2203.04955}, 
}

@article{hafner2019dreamerv1,
  title={Dream to control: Learning behaviors by latent imagination},
  author={Hafner, Danijar and Lillicrap, Timothy and Ba, Jimmy and Norouzi, Mohammad},
  journal={arXiv preprint arXiv:1912.01603},
  year={2019}
}

@article{hafner2020dreamerv2,
  title={Mastering atari with discrete world models},
  author={Hafner, Danijar and Lillicrap, Timothy and Norouzi, Mohammad and Ba, Jimmy},
  journal={arXiv preprint arXiv:2010.02193},
  year={2020}
}

@article{hafner2025dreamerv3,
  title={Mastering diverse control tasks through world models},
  author={Hafner, Danijar and Pasukonis, Jurgis and Ba, Jimmy and Lillicrap, Timothy},
  journal={Nature},
  pages={1--7},
  year={2025},
  publisher={Nature Publishing Group}
}

@article{henaff2017model,
  title={Model-based planning with discrete and continuous actions},
  author={Henaff, Mikael and Whitney, William F and LeCun, Yann},
  journal={arXiv preprint arXiv:1705.07177},
  year={2017}
}

@misc{bengio2015scheduledsamplingsequenceprediction,
      title={Scheduled Sampling for Sequence Prediction with Recurrent Neural Networks}, 
      author={Samy Bengio and Oriol Vinyals and Navdeep Jaitly and Noam Shazeer},
      year={2015},
      eprint={1506.03099},
      archivePrefix={arXiv},
      primaryClass={cs.LG},
      url={https://arxiv.org/abs/1506.03099}, 
}

@misc{asadi2019combatingcompoundingerrorproblemmultistep,
      title={Combating the Compounding-Error Problem with a Multi-step Model}, 
      author={Kavosh Asadi and Dipendra Misra and Seungchan Kim and Michel L. Littman},
      year={2019},
      eprint={1905.13320},
      archivePrefix={arXiv},
      primaryClass={cs.LG},
      url={https://arxiv.org/abs/1905.13320}, 
}

@article{tunyasuvunakool2020,
  title = {dm\_control: Software and tasks for continuous control},
  journal = {Software Impacts},
  volume = {6},
  pages = {100022},
  year = {2020},
  issn = {2665-9638},
  doi = {https://doi.org/10.1016/j.simpa.2020.100022},
  url = {https://www.sciencedirect.com/science/article/pii/S2665963820300099},
  author = {Saran Tunyasuvunakool and Alistair Muldal and Yotam Doron and
            Siqi Liu and Steven Bohez and Josh Merel and Tom Erez and
            Timothy Lillicrap and Nicolas Heess and Yuval Tassa}
}

@article{Bellemare_2013,
   title={The Arcade Learning Environment: An Evaluation Platform for General Agents},
   volume={47},
   ISSN={1076-9757},
   url={http://dx.doi.org/10.1613/jair.3912},
   DOI={10.1613/jair.3912},
   journal={Journal of Artificial Intelligence Research},
   publisher={AI Access Foundation},
   author={Bellemare, M. G. and Naddaf, Y. and Veness, J. and Bowling, M.},
   year={2013},
   month=June, pages={253–279} }

@article{sutton1991dyna,
  title={Dyna, an integrated architecture for learning, planning, and reacting},
  author={Sutton, Richard S},
  journal={ACM Sigart Bulletin},
  volume={2},
  number={4},
  pages={160--163},
  year={1991},
  publisher={ACM New York, NY, USA}
}

@article{lecun2022path,
  title={A path towards autonomous machine intelligence version 0.9. 2, 2022-06-27},
  author={LeCun, Yann},
  journal={Open Review},
  volume={62},
  number={1},
  pages={1--62},
  year={2022}
}

@inproceedings{
    admpo,
    author       = {Haoxin Lin and
                    Yu{-}Yan Xu and
                    Yihao Sun and
                    Zhilong Zhang and
                    Yi{-}Chen Li and
                    Chengxing Jia and
                    Junyin Ye and
                    Jiaji Zhang and
                    Yang Yu},
    title        = {Any-step Dynamics Model Improves Future Predictions for Online and Offline Reinforcement Learning},
    booktitle    = {The 13th International Conference on Learning Representations (ICLR'25)},
    year         = {2025},
    address      = {Singapore}
}

@inproceedings{
    adm2,
    author       = {Haoxin Lin and
                    Siyuan Xiao and
                    Yi-Chen Li and
                    Zhilong Zhang and
                    Yihao Sun and
                    Chengxing Jia and
                    Yang Yu},
    title        = {ADM-v2: Pursuing Full-Horizon Roll-out in Dynamics Models for Offline Policy Learning and Evaluation},
    booktitle    = {The 14th International Conference on Learning Representations (ICLR'26)},
    year         = {2026},
    address      = {Rio de Janeiro, Brazil}
}

@misc{zhang2023leveragingjumpymodelsplanning,
      title={Leveraging Jumpy Models for Planning and Fast Learning in Robotic Domains}, 
      author={Jingwei Zhang and Jost Tobias Springenberg and Arunkumar Byravan and Leonard Hasenclever and Abbas Abdolmaleki and Dushyant Rao and Nicolas Heess and Martin Riedmiller},
      year={2023},
      eprint={2302.12617},
      archivePrefix={arXiv},
      primaryClass={cs.RO},
      url={https://arxiv.org/abs/2302.12617}, 
}

@misc{zhang2026hierarchicalplanninglatentworld,
      title={Hierarchical Planning with Latent World Models}, 
      author={Wancong Zhang and Basile Terver and Artem Zholus and Soham Chitnis and Harsh Sutaria and Mido Assran and Randall Balestriero and Amir Bar and Adrien Bardes and Yann LeCun and Nicolas Ballas},
      year={2026},
      eprint={2604.03208},
      archivePrefix={arXiv},
      primaryClass={cs.LG},
      url={https://arxiv.org/abs/2604.03208}, 
}

@misc{machado2023temporalabstractionreinforcementlearning,
      title={Temporal Abstraction in Reinforcement Learning with the Successor Representation}, 
      author={Marlos C. Machado and Andre Barreto and Doina Precup and Michael Bowling},
      year={2023},
      eprint={2110.05740},
      archivePrefix={arXiv},
      primaryClass={cs.LG},
      url={https://arxiv.org/abs/2110.05740}, 
}

@misc{jayaraman2018timeagnosticpredictionpredictingpredictable,
      title={Time-Agnostic Prediction: Predicting Predictable Video Frames}, 
      author={Dinesh Jayaraman and Frederik Ebert and Alexei A. Efros and Sergey Levine},
      year={2018},
      eprint={1808.07784},
      archivePrefix={arXiv},
      primaryClass={cs.CV},
      url={https://arxiv.org/abs/1808.07784}, 
}

@misc{bruce2024genie,
      title={Genie: Generative Interactive Environments}, 
      author={Jake Bruce and Michael Dennis and Ashley Edwards and Jack Parker-Holder and Yuge Shi and Edward Hughes and Matthew Lai and Aditi Mavalankar and Richie Steigerwald and Chris Apps and Yusuf Aytar and Sarah Bechtle and Feryal Behbahani and Stephanie Chan and Nicolas Heess and Lucy Gonzalez and Simon Osindero and Sherjil Ozair and Scott Reed and Jingwei Zhang and Konrad Zolna and Jeff Clune and Nando de Freitas and Satinder Singh and Tim Rocktäschel},
      year={2024},
      eprint={2402.15391},
      archivePrefix={arXiv},
      primaryClass={cs.LG},
      url={https://arxiv.org/abs/2402.15391}, 
}

@misc{zhao2023learning,
      title={Learning Fine-Grained Bimanual Manipulation with Low-Cost Hardware}, 
      author={Tony Z. Zhao and Vikash Kumar and Sergey Levine and Chelsea Finn},
      year={2023},
      eprint={2304.13705},
      archivePrefix={arXiv},
      primaryClass={cs.RO},
      url={https://arxiv.org/abs/2304.13705}, 
}

@misc{kang2025farvideogenerationworld,
      title={How Far is Video Generation from World Model: A Physical Law Perspective}, 
      author={Bingyi Kang and Yang Yue and Rui Lu and Zhijie Lin and Yang Zhao and Kaixin Wang and Gao Huang and Jiashi Feng},
      year={2025},
      eprint={2411.02385},
      archivePrefix={arXiv},
      primaryClass={cs.CV},
      url={https://arxiv.org/abs/2411.02385}, 
}

@misc{eysenbach2018diversity,
      title={Diversity is All You Need: Learning Skills without a Reward Function}, 
      author={Benjamin Eysenbach and Abhishek Gupta and Julian Ibarz and Sergey Levine},
      year={2018},
      eprint={1802.06070},
      archivePrefix={arXiv},
      primaryClass={cs.AI},
      url={https://arxiv.org/abs/1802.06070}, 
}

@inproceedings{
gumbsch2024learning,
title={Learning Hierarchical World Models with Adaptive Temporal Abstractions from Discrete Latent Dynamics},
author={Christian Gumbsch and Noor Sajid and Georg Martius and Martin V. Butz},
booktitle={The Twelfth International Conference on Learning Representations},
year={2024},
url={https://openreview.net/forum?id=TjCDNssXKU}
}

@misc{vaswani2023attention,
      title={Attention Is All You Need}, 
      author={Ashish Vaswani and Noam Shazeer and Niki Parmar and Jakob Uszkoreit and Llion Jones and Aidan N. Gomez and Lukasz Kaiser and Illia Polosukhin},
      year={2023},
      eprint={1706.03762},
      archivePrefix={arXiv},
      primaryClass={cs.CL},
      url={https://arxiv.org/abs/1706.03762}, 
}

@misc{perez2017film,
      title={FiLM: Visual Reasoning with a General Conditioning Layer}, 
      author={Ethan Perez and Florian Strub and Harm de Vries and Vincent Dumoulin and Aaron Courville},
      year={2017},
      eprint={1709.07871},
      archivePrefix={arXiv},
      primaryClass={cs.CV},
      url={https://arxiv.org/abs/1709.07871}, 
}

@misc{su2023rope,
      title={RoFormer: Enhanced Transformer with Rotary Position Embedding}, 
      author={Jianlin Su and Yu Lu and Shengfeng Pan and Ahmed Murtadha and Bo Wen and Yunfeng Liu},
      year={2023},
      eprint={2104.09864},
      archivePrefix={arXiv},
      primaryClass={cs.CL},
      url={https://arxiv.org/abs/2104.09864}, 
}

@inproceedings{neuralode,
 author = {Chen, Ricky T. Q. and Rubanova, Yulia and Bettencourt, Jesse and Duvenaud, David K},
 booktitle = {Advances in Neural Information Processing Systems},
 editor = {S. Bengio and H. Wallach and H. Larochelle and K. Grauman and N. Cesa-Bianchi and R. Garnett},
 pages = {},
 publisher = {Curran Associates, Inc.},
 title = {Neural Ordinary Differential Equations},
 url = {https://proceedings.neurips.cc/paper_files/paper/2018/file/69386f6bb1dfed68692a24c8686939b9-Paper.pdf},
 volume = {31},
 year = {2018}
}

@misc{talvitie2017selfcorrectingmodelsmodelbasedreinforcement,
      title={Self-Correcting Models for Model-Based Reinforcement Learning}, 
      author={Erik Talvitie},
      year={2017},
      eprint={1612.06018},
      archivePrefix={arXiv},
      primaryClass={cs.LG},
      url={https://arxiv.org/abs/1612.06018}, 
}

@ARTICLE{Bengio94learning,
  author={Bengio, Y. and Simard, P. and Frasconi, P.},
  journal={IEEE Transactions on Neural Networks}, 
  title={Learning long-term dependencies with gradient descent is difficult}, 
  year={1994},
  volume={5},
  number={2},
  pages={157-166},
  doi={10.1109/72.279181}}

@misc{asadi2018lipschitzcontinuitymodelbasedreinforcement,
      title={Lipschitz Continuity in Model-based Reinforcement Learning}, 
      author={Kavosh Asadi and Dipendra Misra and Michael L. Littman},
      year={2018},
      eprint={1804.07193},
      archivePrefix={arXiv},
      primaryClass={cs.LG},
      url={https://arxiv.org/abs/1804.07193}, 
}
